\documentclass[letterpaper]{article} 
\usepackage{aaai2026_camera}  
\usepackage{times}  
\usepackage{helvet}  
\usepackage{courier}  
\usepackage[hyphens]{url}  
\usepackage{graphicx} 
\usepackage{natbib}  
\usepackage{caption} 
\usepackage{url}            
\usepackage{booktabs}       
\usepackage{amsfonts}       
\usepackage{nicefrac}       
\usepackage{microtype}      
\usepackage{xcolor}         
\usepackage{algorithm,algorithmic}
\usepackage{amsmath}
\usepackage{bm}
\usepackage{graphicx}
\usepackage{subcaption}
\usepackage{colortbl,pifont,lipsum,graphics}
\usepackage{multirow}
\usepackage{booktabs}
\usepackage{caption}
\usepackage{makecell}
\usepackage{caption}
\usepackage{pifont}

\def\ie{{\textit{i.e.}}}

\usepackage{arydshln}
\usepackage{algorithm}
\usepackage{algorithmic}

\usepackage{newfloat}
\usepackage{listings}
\DeclareCaptionStyle{ruled}{labelfont=normalfont,labelsep=colon,strut=off} 
\floatstyle{ruled}
\newfloat{listing}{tb}{lst}{}
\floatname{listing}{Listing}
\title{OmniSparse: Training-Aware Fine-Grained Sparse Attention for Long-Video MLLMs}

\author {
    Feng Chen\textsuperscript{\rm 1}$^{\ast}$,
    Yefei He\textsuperscript{\rm 2}$^{\ast}$,
    Shaoxuan He\textsuperscript{\rm 2}$^{\ast}$,
    Yuanyu He\textsuperscript{\rm 2}$^{\ast}$,
    Jing Liu\textsuperscript{\rm 3},
    Lequan Lin\textsuperscript{\rm 4},
    Akide Liu\textsuperscript{\rm 3}, \\
    Zhaoyang Li\textsuperscript{\rm 5},
    Jiyuan Zhang\textsuperscript{\rm 5},
    Zhenbang Sun\textsuperscript{\rm 5},
    Bohan Zhuang\textsuperscript{\rm 2}$^{\ddagger}$,
    Qi Wu\textsuperscript{\rm 1}
}
\affiliations {
    \textsuperscript{\rm 1}AIML, The University of Adelaide, Australia\\
    \textsuperscript{\rm 2}Zhejiang University, China \\ \textsuperscript{\rm 3}Monash University, Australia\\
    \textsuperscript{\rm 4}The University of Sydney, Australia \\
    \textsuperscript{\rm 5}Tiktok, Australia  \\
      \footnotesize{
  $^{\ast}$ Equal contribution \quad
  $\dagger$ Work done during an internship at Tiktok \quad
  $^{\ddagger}$ Corresponding authors
  }
    
}

\usepackage{bibentry}

\begin{document}

\maketitle

\begin{abstract}

Existing sparse attention methods primarily target inference-time acceleration by selecting critical tokens under predefined sparsity patterns. However, they often fail to bridge the training–inference gap and lack the capacity for fine-grained token selection across multiple dimensions—such as queries, key-values (KV), and heads—leading to suboptimal performance and 
acceleration gains.
In this paper, we introduce \texttt{OmniSparse}, a training-aware fine-grained sparse attention of long-video MLLMs, which is applied in both training and inference with dynamic token budget allocation. Specifically, OmniSparse contains three adaptive and complementary mechanisms: (1) query selection as lazy-active classification, aiming to retain active queries that capture broader semantic similarity, while discarding most of lazy ones that focus on limited local context and exhibit high functional redundancy with their neighbors, (2) KV selection with head-level dynamic budget allocation, where a shared budget is determined based on the flattest head and applied uniformly across all heads to ensure attention recall after selection, and (3) KV cache slimming to alleviate head-level redundancy, which selectively fetches visual KV cache according to the head-level decoding query pattern.
Experimental results demonstrate that OmniSparse can achieve comparable performance with full attention, achieving 2.7$\times$ speedup during prefill and 2.4$\times$ memory reduction for decoding.

\end{abstract}
\section{Introduction}
\label{sec:intro}

Long-video multimodal large language models (MLLMs)~\cite{li2025maxinfo,longvlm,longvila} are crucial for understanding complex temporal interactions, but their application is limited by the quadratic computational cost of attention mechanism~\cite{vaswani2017attention}. 
Recent studies~\cite{fastv,mminference} primarily focus on training-free sparse attention, leveraging the inherent sparsity of attention to predefine sparsity patterns, 
to significantly reduce computational overhead during inference. 
For instance, FastV~\cite{fastv} reduces half of the visual tokens after layer 2 based on the attention scores. MMInference~\cite{mminference} introduces a modality-aware dynamic sparse attention mechanism that performs predefined pattern search to accelerate the prefill stage.
These methods achieve significant hardware efficiency but still struggle with training-inference inconsistency and fine-grained token selection across multiple dimensions, resulting in suboptimal performance and limited acceleration gains.

The training-inference gap stems from the use of sparse attention exclusively at inference time, while models are trained under full attention~\cite{chen2025zipr1}. This mismatch leads to inconsistent attention patterns between training and deployment, ultimately degrading generalization and performance. In addition, fine-grained token selection aims to eliminate redundant computations by identifying and removing token-level redundancies across multiple dimensions—namely queries, key-values, and attention heads. However, existing methods typically focus on only one or two of these dimensions, limiting their ability to fully exploit the potential for computational savings and model efficiency.

\begin{figure*}[tb]
  \centering
  \subfloat[Query redundancy in MLLMs.]{
    \label{sfig: query}
    \includegraphics[height=0.54\columnwidth]{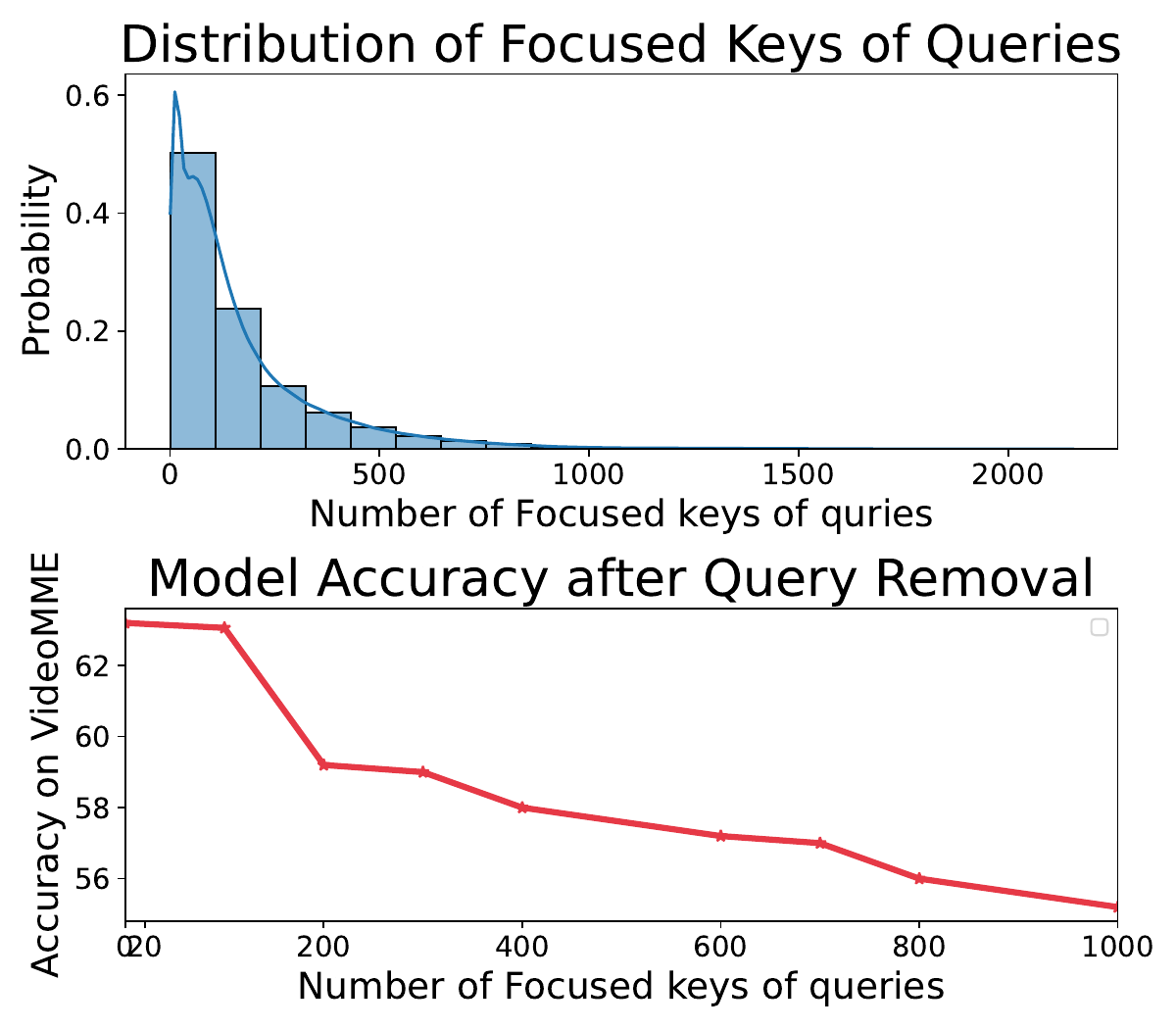}}
    \hspace{0.05em}
  \subfloat[Dynamic sparsity in attention.]{
    \label{sfig: sparsity}
    \includegraphics[height=0.54\columnwidth]{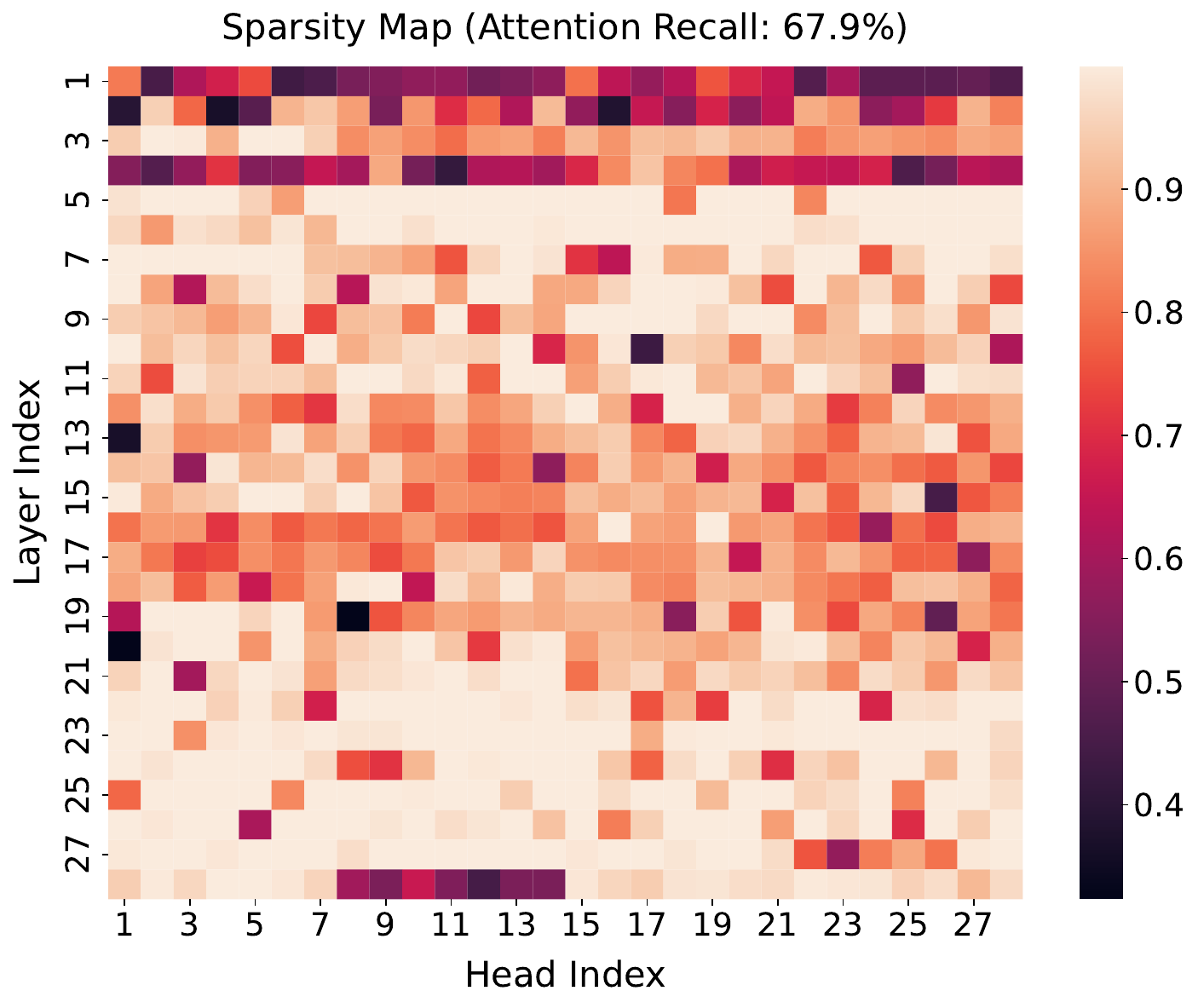}}
  \subfloat[Focus diversity in heads.]{
    \label{sfig: focus}
    \includegraphics[height=0.54\columnwidth]{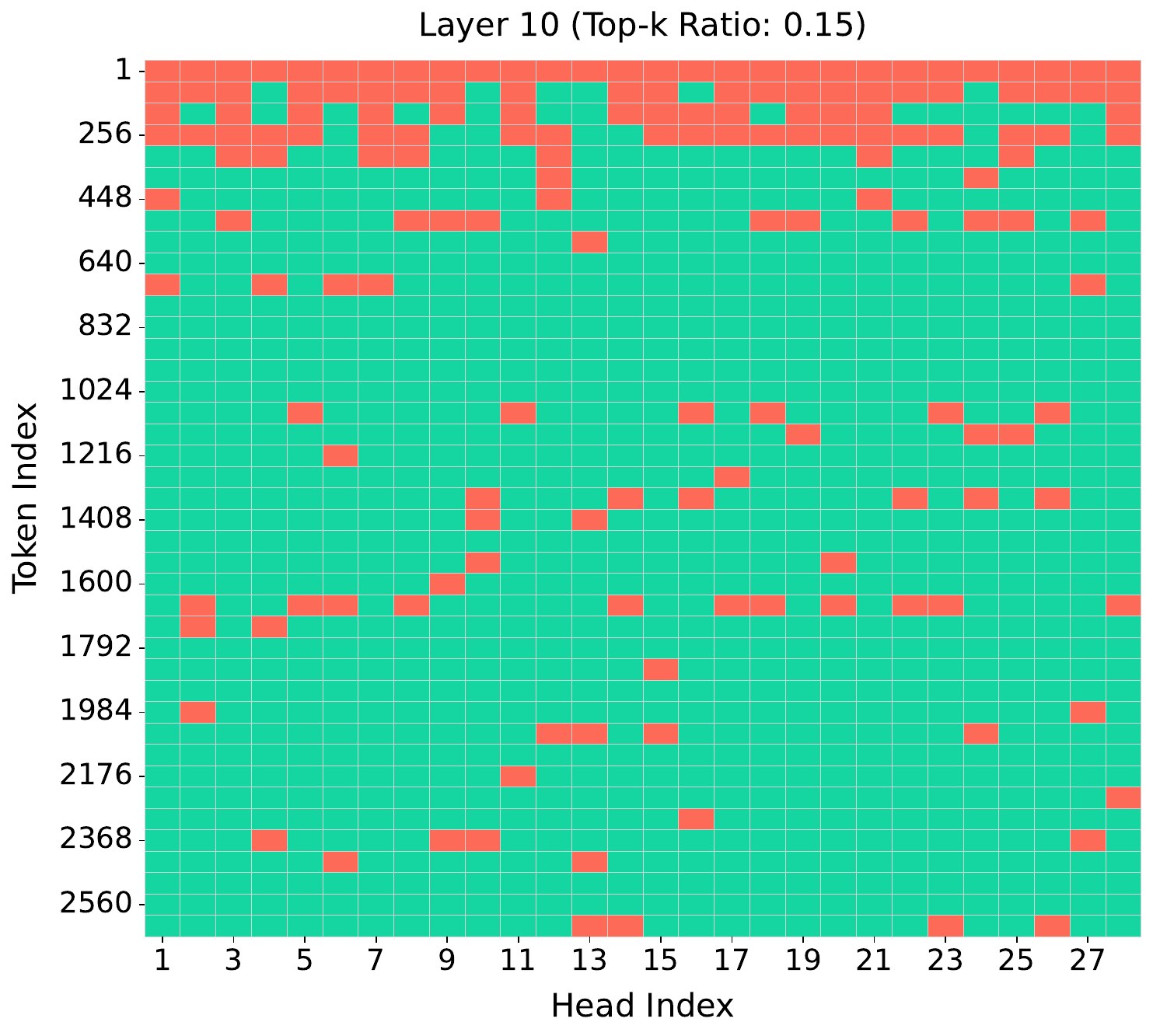}}
  \caption{(a) The majority of queries focus on fewer than 100 out of 2,600 tokens and can be pruned from attention with minimal performance degradation.
(b) The dynamic sparsity across layers and heads suggests that head-wise budget allocation could improve efficiency, but determining an optimal budget for each head is computationally expensive.
(c) Heterogeneous token focus across heads (selected keys are highlighted in red) necessitates the head-level KV selection. Data collected from LLaVA-Video-7b~\cite{llavavideo} with VideoMME~\cite{videomme}.}
\end{figure*}

In this paper, we propose \texttt{OmniSparse}, a training-aware fine-grained sparse attention for long-video MLLMs. OmniSparse dynamically determines token selection budgets across queries, key-values, and attention heads, and applies the same sparse attention during both training and inference to ensure consistency. It adopts a Top-$p$ token-wise sparsification strategy and primarily consists of query selection to remove most functionality-overlapped queries, key-value (KV) selection to determine the minimum token budget across heads, and KV cache slimming to reduce head-wise redundancy. The first two components are designed to alleviate the computation cost of the prefill phase, while the last component focuses on eliminating the memory redundancy of the decoding phase.

\textbf{Query selection as binary lazy-active classification.} Although the query-key angle is designed to capture semantic similarity between tokens~\cite{zhang2024selective}, not all queries contribute equally. As shown in Figure~\ref{sfig: query}, we observe that the majority of queries focus on fewer than 100 out of 2600 tokens (\ie, less than 3\%), typically concentrating on the attention sink~\cite{streamllm} or spatial-temporal adjacent tokens. These “lazy” queries capture limited semantic similarity and can be safely removed from attention with minimal performance degradation, as their functionality is often overlapped with nearby queries or those from other heads (we will demonstrate in Sec. \ref{sec query}).
To this end, we aim to retain “active” queries that capture broader context, while pruning most of the lazy ones that primarily focus on fewer tokens. We formulate query selection as a binary lazy-active classification task. Specifically, we construct an active probe key by aggregating all visual keys, and take the key of attention sink as the lazy probe key, which is used to absorb unnecessary weights from other keys~\cite{streamllm}. For each query, we compute its similarity to these probe keys, where queries exhibiting high similarity with the attention sink are identified as lazy and excluded from subsequent attention computation. To avoid over-pruning of lazy queries, we preserve the queries in the first attention head, which serves as a reliable channel to maintain information integrity.

\textbf{KV selection with head-level dynamic budget allocation.}
As illustrated in Figure~\ref{sfig: sparsity} and~\ref{sfig: focus}, we observe heterogeneous attention patterns across heads: each head attends to distinct KV pairs and exhibits distinct sparsity due to varying attention distribution~\cite{minference}. This disparity causes suboptimal key-value selection under the predefined Top-$k$ strategy~\cite{fastv,seerattention}, leading to under-selection of flatter heads~\cite{twilight}—which have relatively uniform attention weights across tokens—and over-selection of sharp heads.
A straightforward solution is to dynamically allocate KV budgets per head based on their individual attention distributions. However, this fine-grained allocation incurs substantial computational overhead. To address this, we first identify the flattest attention head using kurtosis, as it typically requires a larger token budget to preserve attention recall. Then we allocate the token budget in this head and then uniformly apply it across all heads, allowing each head to select its most salient KV pairs while ensuring the overall attention recall remains above a predefined threshold. 

\textbf{KV cache slimming to alleviate head-wise redundancy.}  Apart from head-wise selection of visual KV pairs, we further reduce memory overhead by conditionally fetching KV caches based on query classification during decoding. Specifically, when a decoding query is identified as lazy, we skip fetching the corresponding head's visual KV cache entirely. This selective fetching strategy significantly decreases the memory access cost during decoding, as it avoids unnecessary KV cache retrievals.

In summary, our contributions are as follows:

(1) We propose \texttt{OmniSparse}, a training-aware fine-grained sparse attention to reduce redundant computation along query, key-value, and head dimensions.

(2) OmniSparse dynamically adapts attention sparsity according to head-specific diversity for efficient prefill, and further reduces memory overhead during decoding by skipping visual KV fetching for heads with lazy decoding queries, enabling fine-grained sparsity across both computation and memory dimensions.

(3) Experimental results demonstrate that OmniSparse attains performance comparable to that of full attention mechanisms, while providing a 2.7$\times$ acceleration during the prefill phase and achieving a 2.4$\times$ reduction in memory consumption during decoding.

\section{Related Work}
\label{sec:formatting}

\begin{figure*}[!t]
    \centering
    \includegraphics[width=0.85\linewidth]{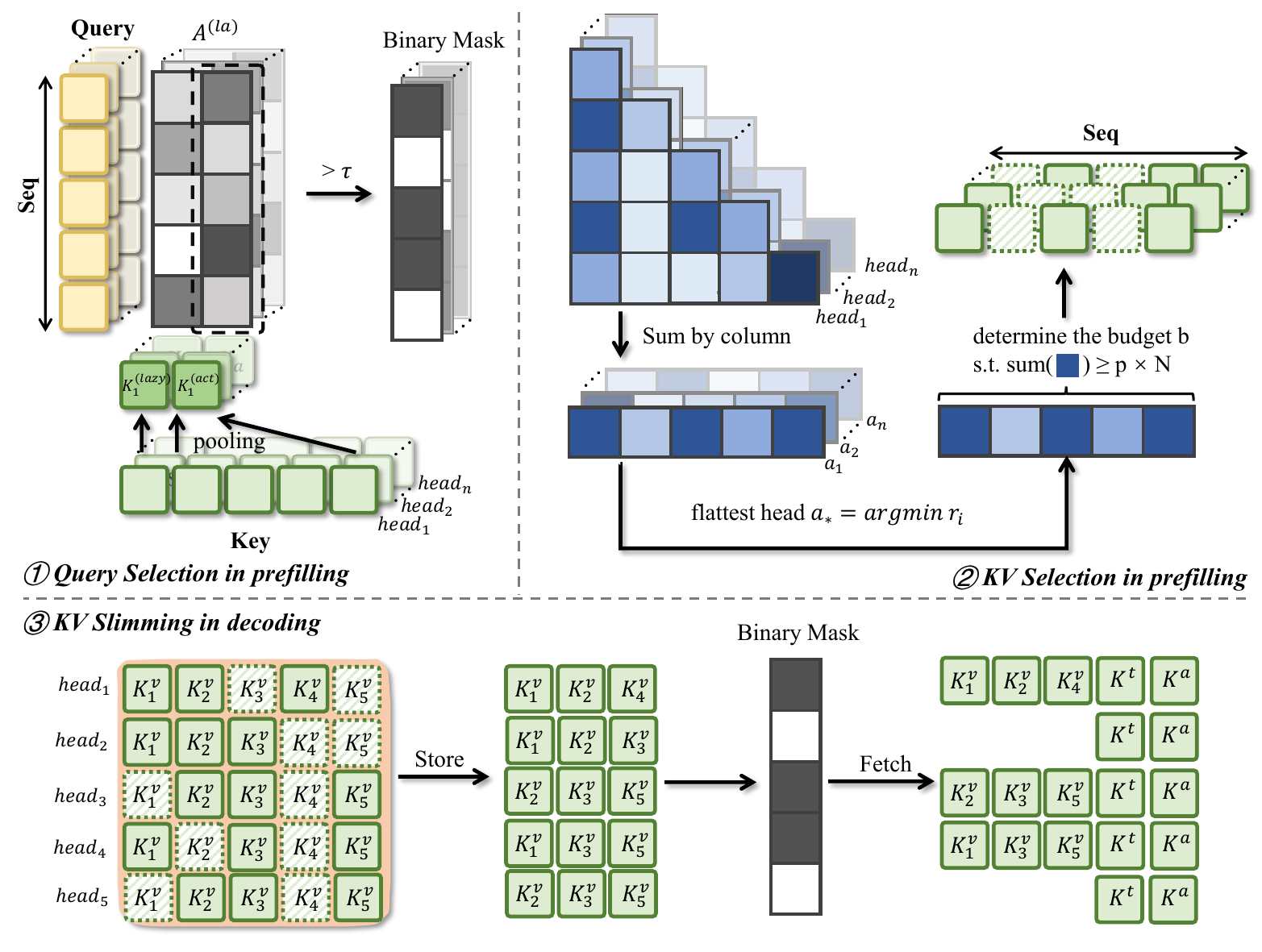}
    \caption{\textbf{Overview of OmniSparse.} During prefill, head-level queries are selected by probing query patterns, with a threshold $\tau$ used to filter out redundant queries. KV selection selects the top $b$ salient KV pairs for each head, with the budget $b$ determined by the flattest head to ensure attention recall exceeds the retention ratio $p$ for all heads. During decoding, only relevant visual KV pairs are fetched for active decoding queries.}
    \label{fig: main}
\end{figure*}

\noindent \textbf{Training-free sparse attention on MLLMs.} Training-free sparse attention is designed to alleviate the computational constraints during deployment, which arise from the extended visual sequence. FastV~\cite{fastv} selects a set of critical tokens after the first layer, reducing the QKV computations of attention to 1/4. FlexPrefill~\cite{lai2025flexprefill} and MMinference~\cite{mminference} dynamically search each head using predefined patterns, accelerating computation with corresponding sparse kernels. VisionZip selects critical visual tokens and merges contextual tokens before the LLM to reduce the number of tokens. AIM~\cite{zhong2024aim} gradually prunes and merges the redundancy tokens based on embedding similarity. However, these methods still struggle with the training-inference gap, leading to an inevitable performance drop and limiting the acceleration gains. 

\noindent \textbf{Long-video MLLMs.} Long-video MLLMs~\cite{longvila,chen2025eagle,shen2024longvu,wang2024longllava} have been developed to tackle the challenges associated with processing prolonged video sequences. One approach focuses on context compression~\cite{longvlm,shen2024longvu}. LongVLM~\cite{longvlm} proposes a hierarchical method to merge local and global information from long-term and short-term video clips, while Maxinfo~\cite{li2025maxinfo} selects key frames and eliminates redundant ones. Another approach extends the context length of LLMs directly~\cite{longvila}. For instance, LongVA~\cite{zhang2024longva} leverages the long-context capabilities of LLMs to process long-video sequences, while LongVITA~\cite{longvita} and LongVILA~\cite{longvila} aim to train models directly on long-video data. However, these models are usually limited by the prolonged input sequence during inference.

\section{Method}
\begin{figure}[!t]
  \centering
  \subfloat{
    \includegraphics[height=0.25\columnwidth]{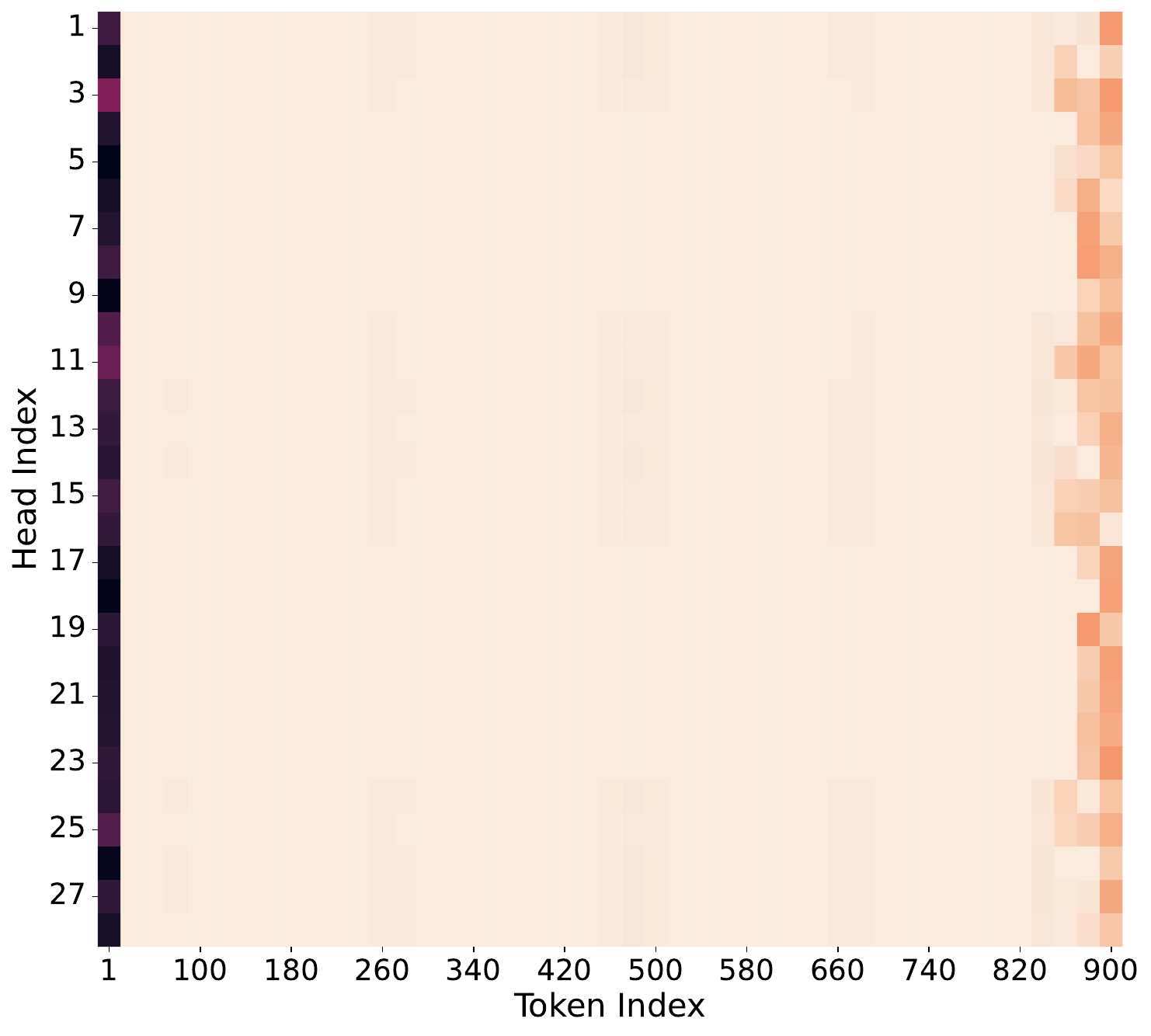}}
    \hspace{0.05em}
  \subfloat{
    \includegraphics[height=0.25\columnwidth]{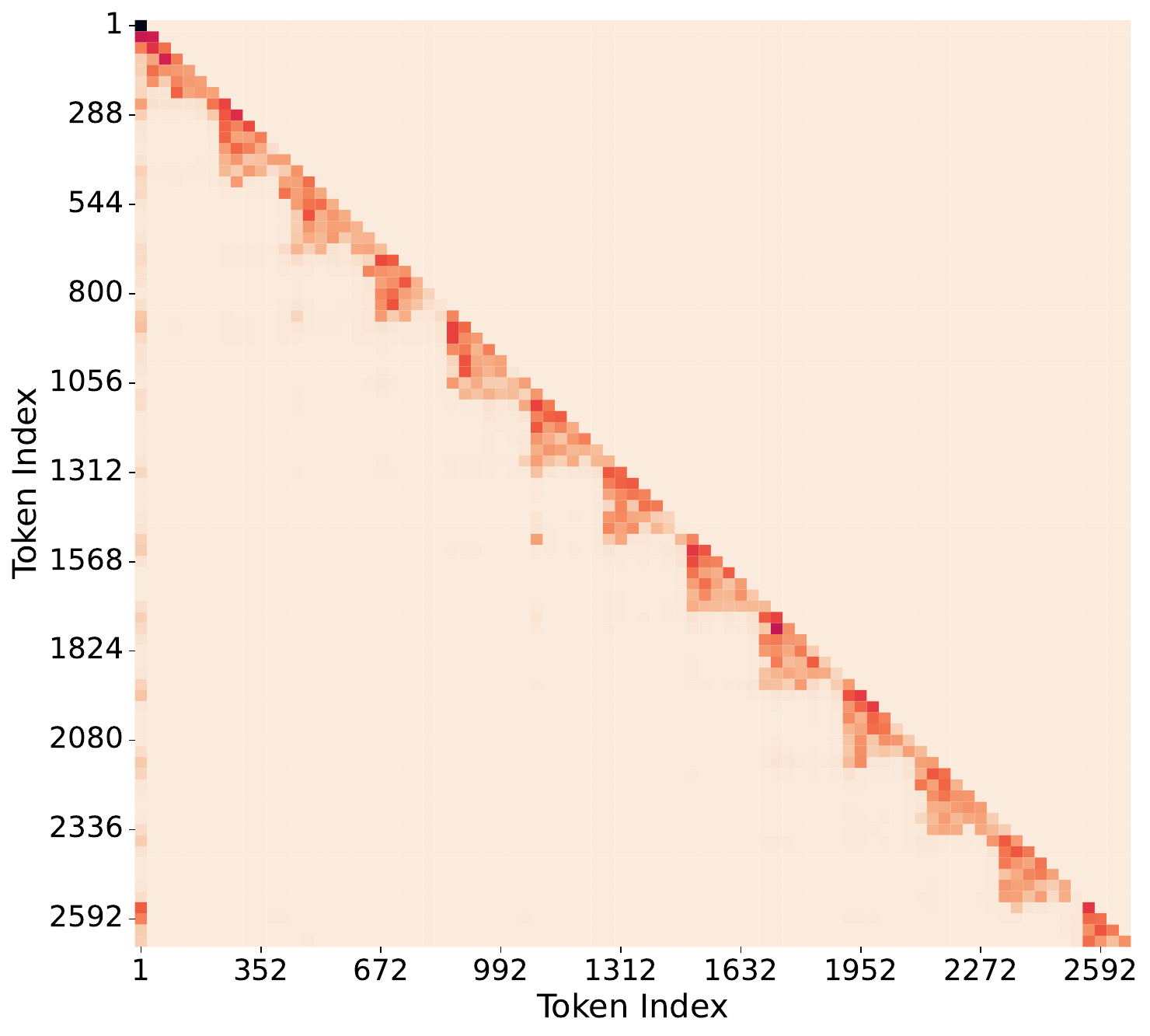}}
  \subfloat{
    \includegraphics[height=0.25\columnwidth]{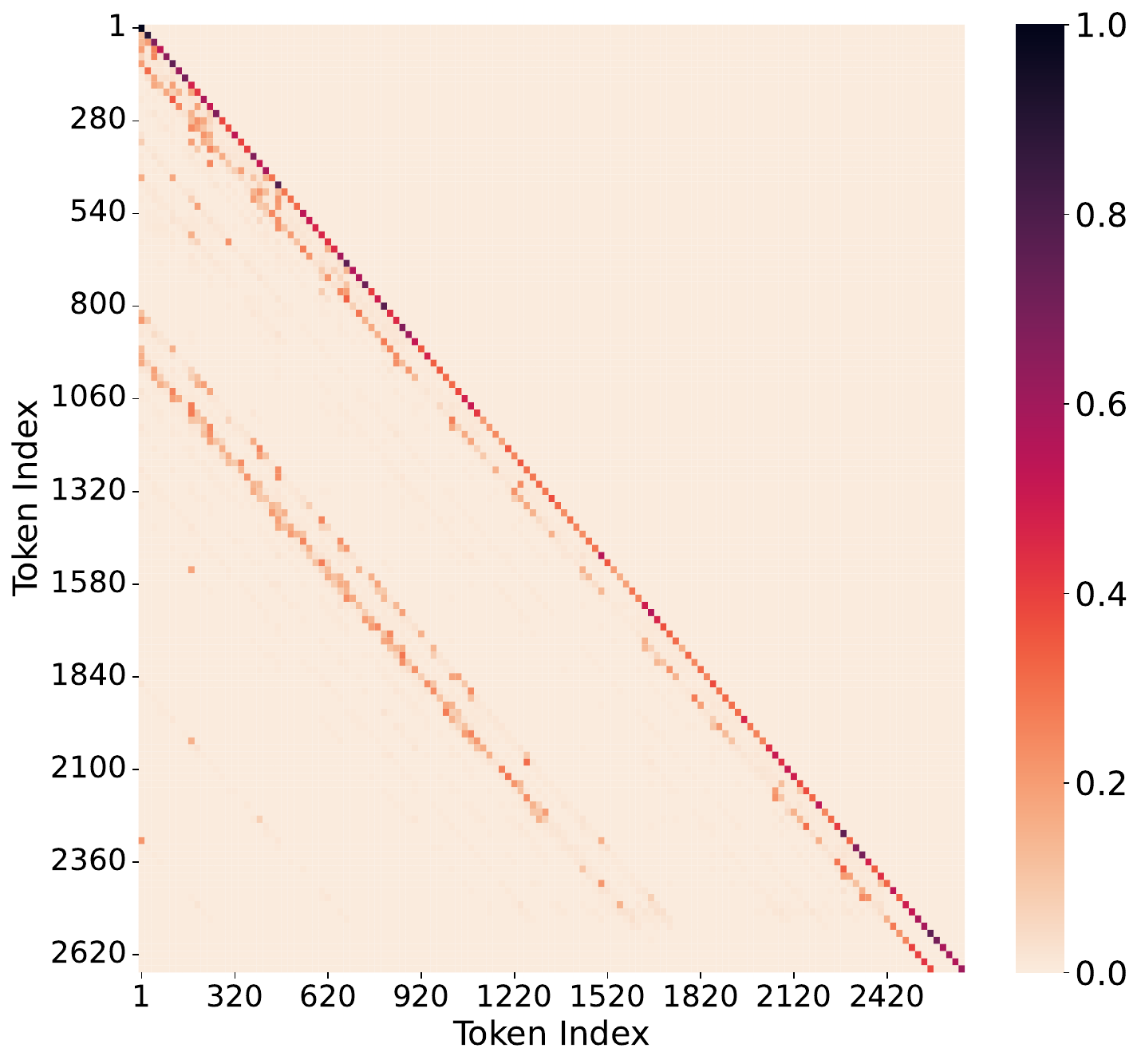}}
    \caption{\textbf{Query redundancy}: Queries from different heads (left), spatially adjacent positions (middle), and temporally adjacent positions (right) focus on similar tokens.}
    \label{fig: overlap}
\end{figure}

\subsection{Preliminary}

\noindent \textbf{Fine-grained token-level sparse attention.} The multi-head attention mechanism enables the model to capture diverse patterns of interactions across different subspaces via learnable linear projections $\mathbf{W}^Q_i, \mathbf{W}^K_i, \mathbf{W}^V _i\in \mathbb{R}^{d \times d_i}$ at each head $i$, where $d_i = d/h$ with $h$ being the number of heads.
Let $\mathbf{Q}_i, \mathbf{K}_i, \mathbf{V}_i$ be the queries, keys, and values at head $i$, respectively. Then, the multi-head attention is mathematically defined as follows:
\begin{align}
\operatorname{MultiHead}(\mathbf{Q}, \mathbf{K}, \mathbf{V}) 
&= \operatorname{Concat}(\text{head}_1, \ldots, \text{head}_h)\mathbf{W}^O, \\
\quad \text{head}_i 
= \mathbf{O}_i = \operatorname{Att}&(\mathbf{Q}_i,\mathbf{K}_i,\mathbf{V}_i) = \sigma\left(\frac{\mathbf{Q}_i\mathbf{K}_i^{\top}}{\sqrt{d_i}}\right)\mathbf{V}_i,
\end{align}
where $\mathrm{Concat}$ denotes horizontal concatenation of matrices, $\mathbf{W}^O \in \mathbb{R}^{d \times d}$ is a linear projection to drive  final output, $\mathbf{O}_i$ is the output of each head, and $\sigma$ is the \texttt{Softmax} activation function. 
In the decoder-only transformer, the self-attention mechanism enables each token in a sequence to attend to preceding tokens when constructing its representation.
Fine-grained token-level sparse attention is to select a subset of most important queries, keys and values $\widetilde{\mathbf{Q}}_i = \mathbf{Q}_i \odot \mathbf{M}^Q_i, \widetilde{\mathbf{K}}_i = \mathbf{K}_i \odot \mathbf{M}^K_i, \widetilde{\mathbf{V}}_i = \mathbf{V}_i \odot \mathbf{M}^V_i$ for each head, where $\mathbf{M}_i^Q, \mathbf{M}_i^K,\mathbf{M}_i^V\in \{0,1\}^{N\times d_i}$ are masks to specify the selection, $N$ is the sequence length, and $\odot$ denotes the element-wise matrix multiplication.

\subsection{Overview}
Given an input sequence $\mathbf{X} = [\mathbf{X}_{v}, \mathbf{X}_{t}]$ comprising language tokens $\mathbf{X}_{t} \in \mathbb{R}^{N_t \times d}$ and vision tokens $\mathbf{X}_{v}\in \mathbb{R}^{N_v \times d}$, where $N_t \ll N_v$ and $N=N_t+N_v$, our proposed \texttt{OmniSparse} module aims to accelerate attention 
by reducing the overhead introduced by the visual modality. As illustrated in Figure \ref{fig: main}, OmniSparse consists of three components: query selection, KV selection, and KV cache slimming. The first two components are designed to accelerate the prefill phase, while the last component aims at speeding up the decoding phase.

\subsection{Query Selection as Binary Lazy-Active Classification}\label{sec query}

 While extensive studies~\cite{moba,zipvl} have explored the sparsity of KV pairs, the redundancy among queries remains under-explored, despite their essential role in capturing token-level semantic similarity~\cite{zhang2024selective}. As shown in Figure~\ref{sfig: query}, we observe that most queries attend to fewer than 100 tokens, and skipping these “lazy” queries during attention leads to minimal performance degradation.
The underlying reason is that the functional role of queries exhibits substantial overlap. A showcase is illustrated in Figure \ref{fig: overlap}, where many queries focus on similar tokens with queries from other heads or neighboring spatial-temporal positions.
Motivated by this observation, we propose to identify and eliminate such lazy queries to reduce redundant computation. We formulate the query pattern as a lazy-active classification problem. Specifically, for each head $i$, we treat the key $\mathbf{K}^{(\text{lazy})}_{i}$ corresponding to the attention sink~\cite{streamllm} (the first token) as the lazy reference, which is a typically lazy pattern and recycles the unnecessary attention weight on broader context~\cite{streamllm}, and compute the active reference key $\mathbf{K}^{(\text{act})}_{i}$ by conducting average pooling over all visual keys. We then construct a compact probe key matrix with these two references and compute binary classification logits as:
\begin{equation}\label{eq:query}
\mathbf{A}^{(\text{la})}_{i} = \sigma \left( \frac{\mathbf{Q}_i \mathbf{K}^{(\text{la})^\top}_{i}}{\sqrt{d_i}} \right), \quad
\mathbf{K}^{(\text{la})}_{i} = \left[\mathbf{K}^{(\text{lazy})}_{i}, \mathbf{K}^{(\text{act})}_{i}\right],
\end{equation}
where $\mathbf{A}^{(\text{la})}_{i} \in \mathbb{R}^{N_v \times 2}$ denotes the attention score over the lazy-active references and serves as classification logits. A query is classified as active if its attention score to the active category exceeds a predefined threshold $\tau$. In addition,  to avoid over-pruning of queries focusing on local context, we preserve the queries in the first attention head as active queries, which serves as a reliable channel to maintain information integrity. Formally, for each head $i$, we construct an active query mask for each token $n$ as:
\begin{equation}\label{eq query2}
\left(\mathbf{M}^Q_i\right)_{n,:} =
\begin{cases}
\mathbf{1}, & \text{if } \left(\mathbf{A}^{(\text{la})}_{i}\right)_{n,2} > \tau  \ \ \text{or} \ \ i=1\\
\mathbf{0}, & \text{otherwise}
\end{cases}
\end{equation}
Then, queries associated with zeros in $\mathbf{M}_i^Q$ are considered lazy and can be skipped in subsequent self-attention layers to avoid redundant computation.







\subsection{KV Selection with Head-level Dynamic Budget Allocation}
While using a predefined global budget for KV selection across all attention heads~\cite{moba} enables efficient batch processing and budget management, it struggles to accommodate the varying sparsity levels among different heads~\cite{zipvl,minference}. To mitigate this limitation, we first identify the flattest attention head, characterized by low token sparsity and typically requiring more tokens to preserve attention recall. We then assign a minimal budget based on this head and use it as a uniform baseline budget for all heads, allowing each head to independently select its most important KV pairs. This approach ensures that, after KV selection, all heads maintain an attention recall above the predefined threshold. Moreover, the method is amenable to efficient implementation via GPU batch processing.
Specifically, for each head $i$, the accumulated attention score for each key $k$ is computed by summing its corresponding row in the attention matrix $\mathbf{A}$: $a_{i,k} = \sum_m (\mathbf{A}_i)_{k,m}$.

To identify the flattest head, we simply evaluate the attention sparsity for each head by computing the kurtosis~\cite{joanes1998comparing} of $a_i$ as $r_i$.
The head with the smallest $r$ is regarded as the flattest, and the KV selection budget is determined based on this head.
Let $a_*$ be the accumulated attention scores at the flattest head. Then we sort the scores in descending order and use \( a_{*}^{\text{sorted}(j)} \) to denote the \( j \)-th highest attention score. We determine the budget \( b \) of important KV pairs by retaining the minimal number of tokens that collectively preserve the majority of the attention weight:
\begin{equation} \label{eq:determine_budget}
b = \min \left\{\, b \in \mathbb{Z} \;\middle|\; \sum_{j=1}^{b} a_{*}^{\text{sorted}(j)} \geq p \times N \,\right\},
\end{equation}
where $N$ is the number of queries, and \( p \) is a threshold that controls the proportion of total attention to retain, which determines a theoretical upper bound of error by $(1-p)\cdot||\mathbf{V}||$~\cite{twilight}. Note that the total sum of attention scores in \( \mathbf{A}_i \) equals \( N \), due to the row-wise \texttt{Softmax} activation. Finally, we select the top \( b \) most critical keys in each head $i$ with the following key mask for each token $n$ as:
\begin{gather} \label{eq:important_token}
\left(\mathbf{M}_i^K\right)_{n, :}  =
\begin{cases}
\mathbf{1}, & \text{if } a_{i,n} \geq a_{i}^{\text{sorted}(b)} \\
\mathbf{0}, & \text{otherwise}
\end{cases}.
\end{gather}
 Since key and value pairs are inherently coupled, we set $\mathbf{M}_i^V=\mathbf{M}_i^K$. Overall, our KV selection strikes a balance between the predefined Top-$k$ strategy across heads~\cite{moba}, which is fixed but efficient, and per-head dynamic allocation, which is adaptive but computationally slower.

\begin{table*}[t]
\centering
\scriptsize{
\begin{tabular}{lccccccc}
\toprule
Model                      & Inference Method & \makecell{Atten FLOPs\\Reduction} & \makecell{KV Cache\\ Reduction} & ActNet-QA & VideoDC & Next-QA & VideoMME \\ \midrule 
\multirow{5}{*}{baseline-256k} & Full            & 0\%                                                                                   & 0\%                                                                                &    57.4               &   3.72              &       79.0          &  63.6              \\
                           & FastV~\cite{fastv}           & 71.7\%                                                                                & 46.4\%                                                                             &  55.8                &   3.68              &    78.4    & 63.3              \\
                           & MInference~\cite{minference}      & 35.3\%                                                                                & 0\%                                                                                &   56.3                &    3.70             &   78.3              & 63.5             \\
                           & ZipVL~\cite{zipvl}           & 63.5\%                                                                                & 40.2\%                                                                             & 56.9          &  3.69                &    78.3             &    63.5           \\
     \rowcolor{gray!15}                          & OmniSparse($\tau=0.08, p=0.82$)   & \textbf{72.6\%}                                                                       & \textbf{53.4\%}                                                                    &  57.4                 &   3.71      &    79.1            & 63.5     \\ \midrule
\multirow{1}{*}{MOBA-256k} & MOBA~\cite{moba}            &   84\%                                                                                 & 0\%                                                                                & 55.4                  &   3.62              &  78.8              & 63.4             \\ \midrule
\multirow{2}{*}{OmniSparse-256k} & OmniSparse($\tau=0.08, p=0.82$)             & 85.7\%                                                                                   & 66.8\%                                                                                & 57.6                  &  3.72               &     78.9            &  63.9             \\
 & OmniSparse($\tau=0.12, p=0.75$)             & 87.1\%                                                                                   & 70.2\%                                                                                & 57.4                   &   3.70               &   78.9              & 63.8              \\  \midrule

 \multirow{2}{*}{OmniSparse-1M} & OmniSparse($\tau=0.08, p=0.82$)             & 86.1\%                                                                                   &  67.1\%                                                                                & 58.2                   &  3.74               &   79.5              & 64.0              \\
 & OmniSparse($\tau=0.12, p=0.75$)             & 87.4\%                                                                                   & 68.6\%                                                                                &  58.1                   &   3.73               &   79.5              & 64.0              \\   \bottomrule

\end{tabular}}\caption{Demonstration of training-aware sparse attention methods across four benchmarks. Models are evaluated with 256 frames and 65,600 tokens. ``OmniSparse-256k" denotes the model trained with OmniSparse and a context length of 256k.}\label{tab: training-aware}
\end{table*}

\textbf{Block-wise probing and sparse flash attention kernel.} To efficiently probe the attention map, we adopt the block-wise probe strategy~\cite{nas,moba,seerattention} via pooling the queries and keys in the sequence dimension to approximate the full attention. 
To avoid the overhead of customizing a probe attention mask for the selected queries from Sec. \ref{sec query}, we instead apply pooling over all queries. Following SeerAttention, we similarly set the block size to 256 and use a customized block-wise sparse Flash-Attention~\cite{dao2023flashattention2} kernel for efficient attention processing.

\subsection{KV Cache Slimming to Alleviate Head-wise Redundancy}

Fetching visual KV caches for all attention heads during decoding introduces redundancy. Since lazy decoding queries contribute little to the generation process, their associated visual KV caches can be omitted to reduce memory access without affecting performance.

Specifically, at each head $i$, let $\mathbf{q}_i \in \mathbb{R}^{d_i}$ be the decoding query, and $\mathbf{K}_i = [\mathbf{K}^{v}_i, \mathbf{K}^{t}_i, \mathbf{K}^{a}_i],\mathbf{V}_i = [\mathbf{V}^{v}_i, \mathbf{V}^{t}_i, \mathbf{V}^{a}_i]$ be the KV cache, where $v, t, a$ denote the KV from vision tokens, text tokens, and the decoded answer, respectively.
First, we store the selected KV pair at the token level. Since each head applies an equal token budget $b$, we can easily index head-wise salient KV pairs in parallel.  
Then, we probe the decoding query pattern in each head using Eq. \ref{eq:query} to identify the lazy decoding query focusing on fewer visual information, then construct the decoding query mask $\mathbf{M}^q_i \in \mathbb{R}^{d_i}$ following Eq. \ref{eq query2}.
We selectively fetch visual KV cache $\mathbf{K}^v_i, \mathbf{V}^v_i$ only for the heads whose query is active (\ie, $\mathbf{M}^q_i =\mathbf{1}$), while skipping visual KV fetches in the heads where $\mathbf{M}^q_i =\mathbf{0}$. Finally, the attention at each head for the decoding token is:
\begin{equation}\label{eq:decoding attention}
\scriptsize{
\mathbf{O}_i =
\sigma\!\left(
    \frac{
        \mathbf{q}_i
        \left[
            \mathbf{K}^v_i \mathrm{diag}(\mathbf{M}^q_i),
            \mathbf{K}^t_i,
            \mathbf{K}^a_i
        \right]^{\!\top}
    }{\sqrt{d_i}}
\right)
\!
\left[
    \mathbf{V}^v_i \mathrm{diag}(\mathbf{M}^q_i),
    \mathbf{V}^t_i,
    \mathbf{V}^a_i
\right].}
\end{equation}

Notably, the head corresponding to a lazy decoding query is not pruned entirely, as it still attends to the KV caches from textual tokens and previously decoded answers.

\section{Experiment}

\begin{table*}[t]
\centering
\scriptsize{
\begin{tabular}{ccccccccc}
\toprule
Model                      & Method & \makecell{Atten FLOPs\\Reduction} & \makecell{KV Cache\\ Reduction} & ActNet-QA & VideoDC & Next-QA & VideoMME & Average\\ \midrule 
\multirow{5}{*}{\makecell{LongVA-7b~\cite{zhang2024longva}\\110 frames \\ 15840 tokens}} & Full            & 0\%                                                                                   & 0\%                                                                                &    \textbf{50.5}                &   \textbf{3.14}               &       67.5          & 52.9      & 50.6        \\
                           & FastV           & 71.7\%                                                                                & 46.4\%                                                                             &     49.7               &   3.06               &    66.9    & 52.0         & 49.8     \\
                           & MInference      & 71.5\%                                                                                & 0\%                                                                                &    49.4                &     3.06             &   67.0              & 52.2      & 49.8        \\
                           & ZipVL           & 79.4\%                                                                                & 55.3\%                                                                             & 50.2          &  3.03                &    67.1             & 52.3     & 50.0         \\
                           & VisionZip           &     78\%                                                                            &  47.0\%                                                                            &  39.6         &   1.06               &    52.4             &   35.2     & 34.5       \\
    \rowcolor{gray!15}                           & OmniSparse(ours)   & \textbf{82.2\%}                                                                       & \textbf{64.9\%}                                                                    &  50.4                  &   3.13      &     \textbf{68.1}            & \textbf{52.9}   & \textbf{50.7}   \\ \midrule
\multirow{5}{*}{\makecell{LLaVA-Video-7b~\cite{llavavideo} \\ 110 frames \\ 20240 tokens}} & Full            & 0\%                                                                                   & 0\%                                                                                & 59.6                   &   3.66               &   81.2              & 64.7   & 60.5           \\
                           & FastV           & 71.7\%                                                                                & 46.4\%                                                                             & 59.2                   &     3.60             &  80.2      & 64.1    & 59.9          \\
                           & MInference      & 22.8\%                                                                                & 0\%                                                                                &  59.6                  &    3.64              &   80.6              & 64.6      &  60.3        \\
                           & ZipVL           & 75.9\%                                                                                & 50.7\%                                                                             & 59.4          & \textbf{3.66}                 &    80.6             & 64.5      &  60.2       \\
                           & VisionZip           &  64.8\%                                                                               &   40.6\%                                                                           & 42.1          &    1.35              &   69.5              &     44.9   & 42.5       \\
     \rowcolor{gray!15}                          & OmniSparse(ours)   & \textbf{75.9\%}                                                                       & \textbf{63.7\%}                                                                    &  \textbf{60.4}                   & 3.65        &   \textbf{81.3}              & \textbf{64.7}   &  \textbf{60.8}  \\ \midrule
\multirow{5}{*}{\makecell{LongVILA-7b~\cite{longvila} \\ 256 frames \\ 65800 tokens}} & Full            & 0\%                                                                                   & 0\%                                                                                &   59.5                 &  2.76                &     80.7            & \textbf{60.1}  & 56.9            \\
                           & FastV           & 71.7\%                                                                                & 46.4\%                                                                             &     59.1               &   2.72               &  80.1      & 57.8     & 56.1         \\
                           & MInference      & 53\%                                                                                & 0\%                                                                                &    \textbf{59.7}                &   2.77               &    79.1             & 60.0   &   56.6        \\
                           & ZipVL           & 82.1\%                                                                                & 57.7\%                                                                             & 57.8          &  2.75                 &    79.5             & 59.4       & 56.0       \\
                           & VisionZip           &                                                                                 &                                                                              &          &      OOM            &                 &               \\
  \rowcolor{gray!15}                         & OmniSparse(ours)   & \textbf{82.3\%}                                                                       & \textbf{68.4\%}                                                                    &  59.6                  & \textbf{2.78}        &   \textbf{80.7}              & 60.0   & \textbf{57.0}  \\ \bottomrule
\end{tabular}}\caption{Inference efficiency comparison of sparse attention methods on four benchmarks.}\label{tab: inference}
\end{table*}

\noindent \textbf{Models.} For training-aware sparse attention comparison, we implement our OmniSparse on LLaVA-Video~\cite{llavavideo} to ensure training-inference consistency, supporting 256k and 1M token length for long video training. We use Long-VITA~\cite{longvita} training data to gradually extend the context from 32k to 256k and then 1M. We adopt Qwen2.5-7b-Instruct~\cite{qwen2.5} as LLM backbone, SigLip-400M~\cite{siglip} as visual encoder, and a 2-layer MLP as the adapter, where each frame is encoded to 256 tokens. The whole training is conducted using 256 H100 GPUs. For training-free comparison, we use LongVA-7b~\cite{zhang2024longva}, LLaVA-Video-7b-Qwen2~\cite{llavavideo} and LongVILA-7b-Qwen2-1M~\cite{longvila} for their proficiency in handling long-context tasks and compare our method with state-of-the-art sparse attention, including  FastV~\cite{fastv}, MInference~\cite{minference}, ZipVL~\cite{zipvl}, and VisionZip~\cite{yang2024visionzip}.

\noindent \textbf{Benchmarks.} To verify the effectiveness, we use ActivityNet-QA~\cite{activitynet}, EgoSchema~\cite{egoschema}, EventBench~\cite{eventbench},  VideoMME~\cite{videomme}, PerceptionTest~\cite{perception}, NExT-QA~\cite{nextqa}, LongVideoBench~\cite{longvideobench}, MVBench~\cite{mvbench}, VNBench~\cite{vnbench}, and VideoDC~\cite{videodc} for video evaluation.

\textbf{Baseline.} For both training-aware and training-free comparison, we treat the model trained with full attention as the baseline and empirically set the hyperparameters of OmniSparse as  $\tau = 0.08$ and $p = 0.82$ for training and inference. Besides, we apply MOBA~\cite{moba} to the baseline as a training-aware sparse attention mechanism for comparison, using a block size of 2048 and selecting the top 20 most important blocks for attention.

\subsection{Performance Comparison}

\noindent \textbf{Training-aware comparison.} We first demonstrate the advantages of training-aware sparse attention, as shown in Table \ref{tab: training-aware}. Results across four video benchmarks indicate that our OmniSparse achieves performance comparable to full-attention training while easily scaling to 1 million contexts. Besides, our method significantly outperforms MOBA~\cite{moba}, a training-aware sparse attention method with head-wise Top-$k$ KV selection strategy. For example, OmniSparse surpasses MOBA by 2.2\% on ActivityNet-QA and provides additional KV cache compression to accelerate the decoding phase. Furthermore, we compare training-aware sparse attention with its training-free counterpart. For our OminiSparse, maintaining consistency between training and inference results in an additional 13.1\% reduction in FLOPs and 13.4\% reduction in memory usage, while still delivering better performance than the training-free approach.

\textbf{Training-free comparison.} We also compare our method with existing sparse attention approaches on mainstream MLLMs for inference acceleration. As shown in Table~\ref{tab: inference}, our method, OmniSparse, not only achieves performance comparable to full attention but also surpasses other sparse attention methods in both FLOPs reduction and KV cache efficiency. For example, compared to the Top-$k$-based FastV~\cite{fastv}, OmniSparse further reduces computation by 4.2\% in FLOPs and memory usage by 17.3\% within the LLaVA-Video~\cite{llavavideo} framework. Relative to the Top-$p$-based ZipVL~\cite{zipvl}, our method enables finer-grained token selection across both query and head dimensions, yielding an additional 16\% reduction in KV cache size on LongVA~\cite{zhang2024longva}. These results suggest that dynamic budget allocation across multiple dimensions offers greater potential than coarse-grained or fixed-budget approaches.
We also present a decoding speed comparison in Table~\ref{tab decoding}, demonstrating that our method outperforms existing approaches in deployment efficiency.

\begin{table}[]
\scriptsize{
\begin{tabular}{cccccc}
\toprule
           & FastV & Minference & ZipVL & VisionZip & OmniSparse \\ \midrule
TTFT (s)      &  13.6     &     12.9       &   13.3    &    11.6       &       10.1     \\
\makecell{Throughput\\(tokens/s)} &    8.5   &    4.3        &   8.4    &    9.7       &      11.1      \\ \bottomrule
\end{tabular}}\caption{Decoding speed comparison with existing sparse attention with a context length of 64k. “TTFT” denotes time-to-first-token and is measured with a batch size of 1 on an Nvidia H100 GPU.}\label{tab decoding}
\end{table}

\begin{figure}[!t]
    \centering
    \includegraphics[width=0.9\linewidth]{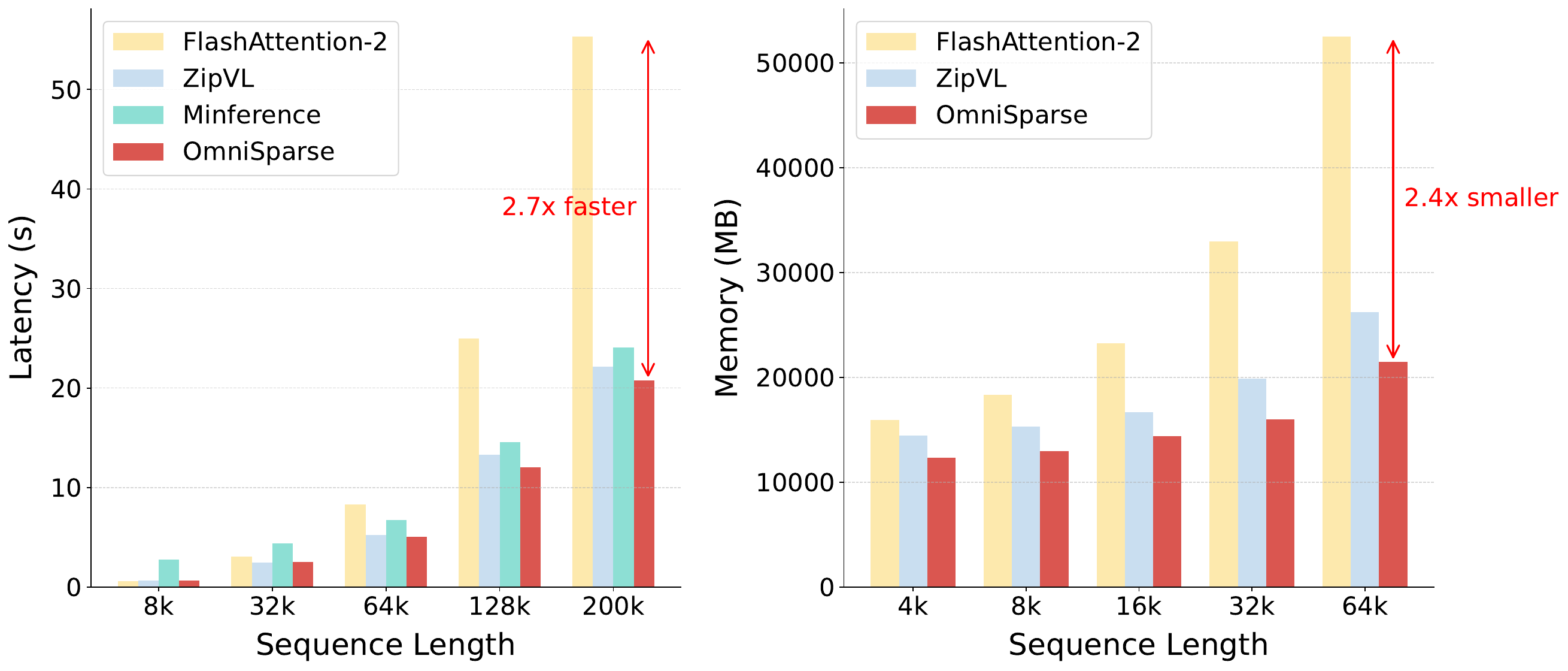}
    \captionof{figure}{Prefill latency (left) and decoding memory usage (right) under varying sequence lengths on LLaVA-Video-7b.}
    \label{fig:latency_memory}
\end{figure}

\noindent    \textbf{Runtime deployment speed.} We present the time-to-first-token (TTFT) and decoding throughput in Table~\ref{tab speed}, comparing the performance of our method against FlashAttention~\cite{dao2023flashattention2} on different input lengths using LLaVA-Video-7b. Our method consistently demonstrates lower latency and higher throughput across all tested input lengths, highlighting its efficiency over FlashAttention, particularly for smaller input sizes. At larger input lengths of 64k, FlashAttention encounters out-of-memory (OOM) errors, while our method remains scalable. Furthermore, we evaluate OmniSparse under varying sequence lengths in terms of latency and memory, as shown in Figure \ref{fig:latency_memory}. Our method achieves a 2.7$\times$ speedup during the prefill stage and a 2.4$\times$ reduction in memory usage during decoding. 

\begin{table}[]
    \centering
\footnotesize{
\begin{tabular}{cccc}
\toprule
Input Length      & Method         & TTFT(s) & \makecell{Throughput\\(tokens/s)} \\ \midrule
\multirow{2}{*}{16k} & FlashAttention &     3.52    &    15.50        \\
                  &   Ours             &   3.02      &   40.61         \\\hdashline
\multirow{2}{*}{32k} & FlashAttention &     6.40    &    OOM        \\
                  &    Ours            &    5.37     &  16.32          \\\hdashline
\multirow{2}{*}{64k} & FlashAttention &    15.45     &    OOM        \\
                  &      Ours          &   10.06      &    11.10        \\\hdashline
\multirow{2}{*}{128k} & FlashAttention &    44.82     &     OOM       \\
                  &     Ours           &   20.05      &  OOM           \\ \bottomrule
\end{tabular}}\captionof{table}{Latency and throughput under different input lengths on LLaVA-Video-7b. “TTFT” denotes time-to-first-token and is measured with a batch size of 1 on an Nvidia H100 GPU.}\label{tab speed}
\end{table}

\subsection{Ablation Study and Discussion}
\noindent \textbf{Ablation on fine-grained selection.} 
 As shown in Table \ref{tab ablation selection}, we conduct a comprehensive ablation study on the effects of query selection, KV selection, and KV cache compression strategies on LLaVA-Video-7b. In the top section, we observe that applying query selection or KV selection individually yields a significant reduction in attention FLOPs (54.4\% and 51.5\%, respectively) without any loss in VideoMME accuracy. When both are applied simultaneously, the computation cost is further reduced (77.9\% reduction), demonstrating the complementary nature of the two strategies. In the bottom section, we evaluate the impact of KV cache compression. While individual techniques such as KV pruning or regrouping provide moderate cache savings (29.7\% and 51.5\%, respectively), combining them leads to substantial memory reduction (64.1\%) with no drop in accuracy. These results confirm that our multi-dimensional sparse attention and KV compression techniques can significantly improve efficiency while preserving performance. The results suggest that multi-dimensional sparsification—spanning both query and KV token selection as well as KV cache compression—enables substantial savings in computation and memory without compromising model performance. Moreover, regarding the first-head query, removing it leads to a noticeable 0.4\% drop in VideoMME accuracy while offering only a marginal additional FLOPs reduction of 3\%. We therefore retain this head to avoid over-pruning queries that capture complementary contextual cues.

\begin{table}[]
    \centering
\tiny{
\begin{tabular}{ccccc}
\toprule
\multicolumn{5}{c}{\textbf{Sparse Attention}} \\ \midrule
Query Selection & KV Seletion & Ratio & Attn FLOPs Reduction & VideoMME \\
                &             &  100\%     &     0\%                 & 64.7         \\
\ding{51}       &             &   72.8\%    &   54.4\%                   &    64.7      \\
                & \ding{51}   &  74.2\%     &      51.5\%                &   64.7       \\
\ding{51}       & \ding{51}   &    47.1\%   &       77.9\%               &  64.7        \\ \midrule
\multicolumn{5}{c}{\textbf{KV Cache Compression}} \\ \midrule
KV Selection      & KV Pruning & KV Regrouping & KV Cache Reduction & VideoMME \\
                &               &      &     0\%               &     64.7     \\
\ding{51}       &               &       &    15.5\%                &   64.7       \\
  \ding{51}               & \ding{51}     &       &       29.7\%             &     64.7     \\
\ding{51}       &      &  \ding{51}     &       51.5\%             &   64.7       \\ 
\ding{51}       & \ding{51}     &  \ding{51}     &    64.1\%                &    64.7      \\ \bottomrule
\end{tabular}} \caption{Ablation of sparse attention and KV cache compression on LLaVA-Video-7b. “Ratio” denotes the proportion of tokens participating in attention computation.}\label{tab ablation selection}
\end{table}

\begin{figure}[h]
    \centering

    \includegraphics[width=0.65\linewidth]{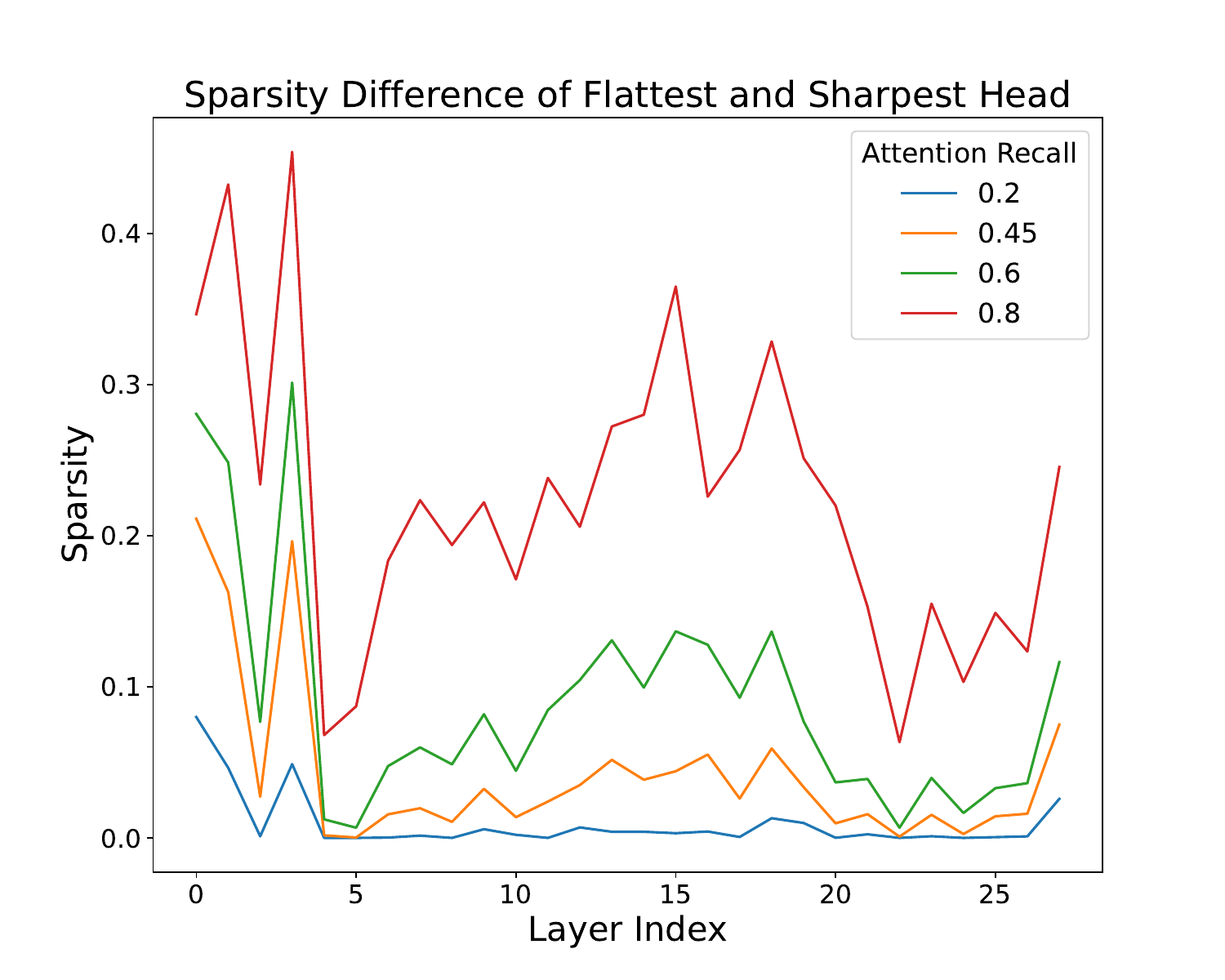}
    \caption{Layer-wise sparsity difference between flattest and sharpest heads on LLaVA-Video-7b.
    }
    \label{fig: sparsity diff}
\end{figure}

\noindent \textbf{Over-selecting KV pairs for sharp heads.} In KV selection of our method, applying the token budget determined by the flattest head to all other heads inevitably leads to over-selection for sharp heads. As shown in Figure~\ref{fig: sparsity diff}, we report the layer-wise sparsity difference between the flattest and sharpest heads. We observe that this over-selection redundancy largely depends on the target attention recall. When the attention recall is set to 0.2, only an additional 5\% of queries are selected, whereas at 0.8, the over-selection increases to approximately 20\%. This sparsity gap tends to grow as the attention recall increases. Therefore, we set $p=0.82$, a moderate attentional recall to balance the performance and efficiency. Moreover, compared to assigning separate token budgets for each head, such redundancy remains within an acceptable range for computation latency.

\section{Conclusion}
In this paper, we have proposed OmniSparse, a training-aware, fine-grained sparse attention framework designed to accelerate inference for long-video MLLMs. OmniSparse keeps training-inference consistency by applying in both training and inference, enabling it to not only approximate the results of full attention but also achieve high acceleration gains. It features multi-dimensional token selection across queries, key-values, and attention heads, and supports optimization for both the prefill and decoding stages. Experimental results have demonstrated that OmniSparse achieves performance comparable to full attention, while delivering a 2.7$\times$ speedup during prefill and a 2.4$\times$ reduction in memory usage during decoding.

\noindent \textbf{Limitations and future work.} While our OmniSparse framework demonstrates promising results in terms of reducing computational overhead and memory usage, several limitations remain. The threshold for lazy-active query classification may vary across layers. This variation could affect the balance between computational efficiency and model performance, particularly in layers where the attention patterns are more dynamic or complex. To address this issue, we plan to further investigate the role of attention layers in video perception and understanding.


\bibliography{LaTeX/main}


\end{document}